\documentclass[11pt,a4paper]{article}
\usepackage[hyperref]{emnlp2020}
\usepackage{times}
\usepackage{latexsym}

\usepackage{microtype}

\usepackage{url}
\usepackage{graphicx}
\usepackage{subcaption}
\usepackage{amsfonts}
\usepackage{multirow}
\usepackage{makecell}

\usepackage{mathrsfs}

\usepackage{amsmath,amssymb}

\aclfinalcopy

\title{WeChat Neural Machine Translation Systems for WMT20}

\author{Fandong Meng, Jianhao Yan, Yijin Liu, Yuan Gao, Xianfeng Zeng, \\ \textbf{Qinsong Zeng, Peng Li, Ming Chen, Jie Zhou, Sifan Liu and Hao Zhou} \\
WeChat AI, Tencent, China\\
{\{fandongmeng,elliottyan,yijinliu,masongao,xianfzeng\}@tencent.com} \\
{\{qinzzeng,patrickpli,mingchen,withtomzhou,stephenliu,harveyzhou\}@tencent.com}
}

\date{}

\begin{document}
\maketitle
\begin{abstract}
We participate in the WMT 2020 shared news translation task on Chinese$\to$English. Our system is based on the Transformer~\cite{transformer2017} with effective variants and the DTMT~\cite{meng2019dtmt} architecture. In our experiments, we employ data selection, several synthetic data generation approaches (i.e., back-translation, knowledge distillation, and iterative in-domain knowledge transfer), 
advanced finetuning approaches and self-bleu based model ensemble. Our constrained Chinese$\to$English system achieves 36.9 case-sensitive BLEU score, which is the highest among all submissions.
\end{abstract}

\section{Introduction} \label{introduction}
Our WeChat AI team participates in the WMT 2020 shared news translation  task on Chinese$\to$English. In this year’s translation task, we mainly focus on exploiting several effective model architectures, better data augmentation, training and model ensemble strategies.

For model architectures, we mainly exploit two different architectures in our approaches, namely Transformers and RNMT. For Transformers, we implement the Deeper transformer with Pre-Norm, the Wider Transformer with larger filter-size and the average attention based transformer~\cite{zhangetalaan2018}.
For the RNMT, we use the deep transition based DTMT~\cite{meng2019dtmt} model. We finally ensemble four kinds of models in our system.

For synthetic data generation, we explore various methods for out-of-domain and in-domain data generation. For out-of-domain data generation, we explore the back-translation method~\cite{sennrichetalimproving2016} to leverage the target side monolingual data and the knowledge distillation method~\cite{kim2016sequence} to leverage source side of golden parallel data.
For in-domain data generation, we employ iterative in-domain knowledge transfer to leverage the source side monolingual data and golden parallel data. 
Furthermore, data augmentation methods, including noisy fake data~\cite{wuetalexploiting2019} and sampling~\cite{edunov2018understanding}, are used for training more robust NMT models. 

For training strategies, we mainly focus on the parallel scheduled sampling~\cite{mihaylovamartinsscheduled2019,parallel2019schedule}, the target denoising and minimum risk training~\cite{shenetalmrt2016,wangsennrichexposure2020} algorithm for in-domain finetuning.

We also exploit a self-bleu \cite{selfbleu2018texygen} based model ensemble approach to enhance our system. As a result, our constrained Chinese$\to$English system achieves the highest case-sensitive BLEU score among all submitted systems.

In the remainder of this paper, we start with an overview of model architectures in Section~\ref{architectures}. 
Section~\ref{techniques} describes the details of our systems and training strategies. 
Then Section~\ref{experiments} shows the experimental settings and results. 
Finally, we conclude our work in Section~\ref{conclusion}.

\section{Model Architectures} \label{architectures}

In this section, we first describe the model architectures we use in the Chinese$\xrightarrow{}$English Shared Task, including the Transformer-based \cite{transformer2017} models and RNN-based \cite{bahdanau2014neural,meng2019dtmt} models.

\subsection{Deeper Transformer}

As shown in previous studies \cite{wang2019learning,baiduwmt2019}, deeper Transformers with pre-norm outperform its shallow counterparts on various machine translation benchmarks. 
In their work, increasing the encoder depth significantly improves the model performance, while they only introduces mild overhead in terms of speed in training and inference, compared with increasing the decoder side depth. 

Hence, we train deeper Transformers with a deep encoder aiming for a better encoding representation. In our experiments, we mainly adopt two settings, with the hidden size 512 (Base) and 1024 (Large).
We adopt a 30-layer encoder for Base models, and 20/24-layer encoders for Large models. 
Further increasing the encoder depth does not lead to a significant BLEU improvement.
To keep the total trainable parameters the same among models, the filter sizes of Base and Large models are 16384 and 4096, respectively.
For training, the batch size is 4,096 tokens per GPU, and we train each model using 8 NVIDIA V100 GPUs for about 7 days.

\subsection{Wider Transformer}

Inspired by last year's Baidu system \cite{baiduwmt2019}, we also train Wider Transformers with larger inner dimension of the Feed-Forward Network than the standard Transformer Large system. 
Specifically, two settings are used in our experiments. 
With filter size as 15,000, we set the number of encoder layers to 10, and with filter size 12,288, we set the number of encoder layers to 12. The number of total trainable parameters of Wider Transformer is kept approximately the same as our Deeper Transformers.

In our experiments, we also set the batch size to be 4,096 and train the Wider Transformers with 8 NVIDIA V100 GPUs for about 7 days. 

\subsection{Average Attention Transformer}
To introduce more diversity in our Transformer models, we use Average Attention Transformer (AAN) \cite{zhangetalaan2018} as one of our candidate architectures. The Average Attention Transformer replaces the decoder self-attention module in auto-regressive order with a simple average attention, and introduces almost no loss in model performance.

We believe that even though the performance of AAN does not drop in terms of BLEU, the output distributions of AAN networks should be different from the output distributions of original Transformers, which brings diversity for final ensemble.
This also complies with our findings in self-bleu experiments (Section \ref{sec:ens}). 

In practice, AAN models are trained for both the Wider Transformer and Deeper Transformer. The batch size and other hyper-parameters are kept the same its non-AAN counterpart.

\subsection{DTMT}

DTMT~\cite{meng2019dtmt} is the recently proposed deep transition RNN-based model for Neural Machine Translation, whose encoder and decoder are composed of the well designed transition blocks, each of which consists of a
linear transformation enhanced GRU (L-GRU) followed by several transition GRUs (T-GRUs). 
DTMT enhances the hidden-to-hidden transition with multiple non-linear transformations, as well as maintains a linear transformation path throughout this deep transition by the well-designed linear transformation mechanism to alleviate the vanishing gradient problem.
This architecture has demonstrated its superiority over the conventional Transformer model and stacked RNN-based models in NMT~\cite{meng2019dtmt}, and also achieves surprising performances on other NLP tasks, such as sequence labeling~\cite{liuetalgcdt2019} and aspect-based sentiment analysis~\cite{liangetalabsa2019}.

In our experiments, we use the bidirectional deep transition encoder, where each directional deep transition block consists of 1 L-GRU and 4 T-GRU. 
The decoder contains a query transition block and the decoder transition block,  each of which consists of 1 L-GRU and 4 T-GRU. 
Therefore the DTMT consists of a 5 layer encoder and a 10 layer decoder, with hidden size 1,024. 
We use 8 NVIDIA V100 GPUs to train each model for about three weeks and the batch size is set to 4,096 tokens per GPU. 

\begin{figure*}[t!]
\begin{center}
      \includegraphics[width=0.8\textwidth]{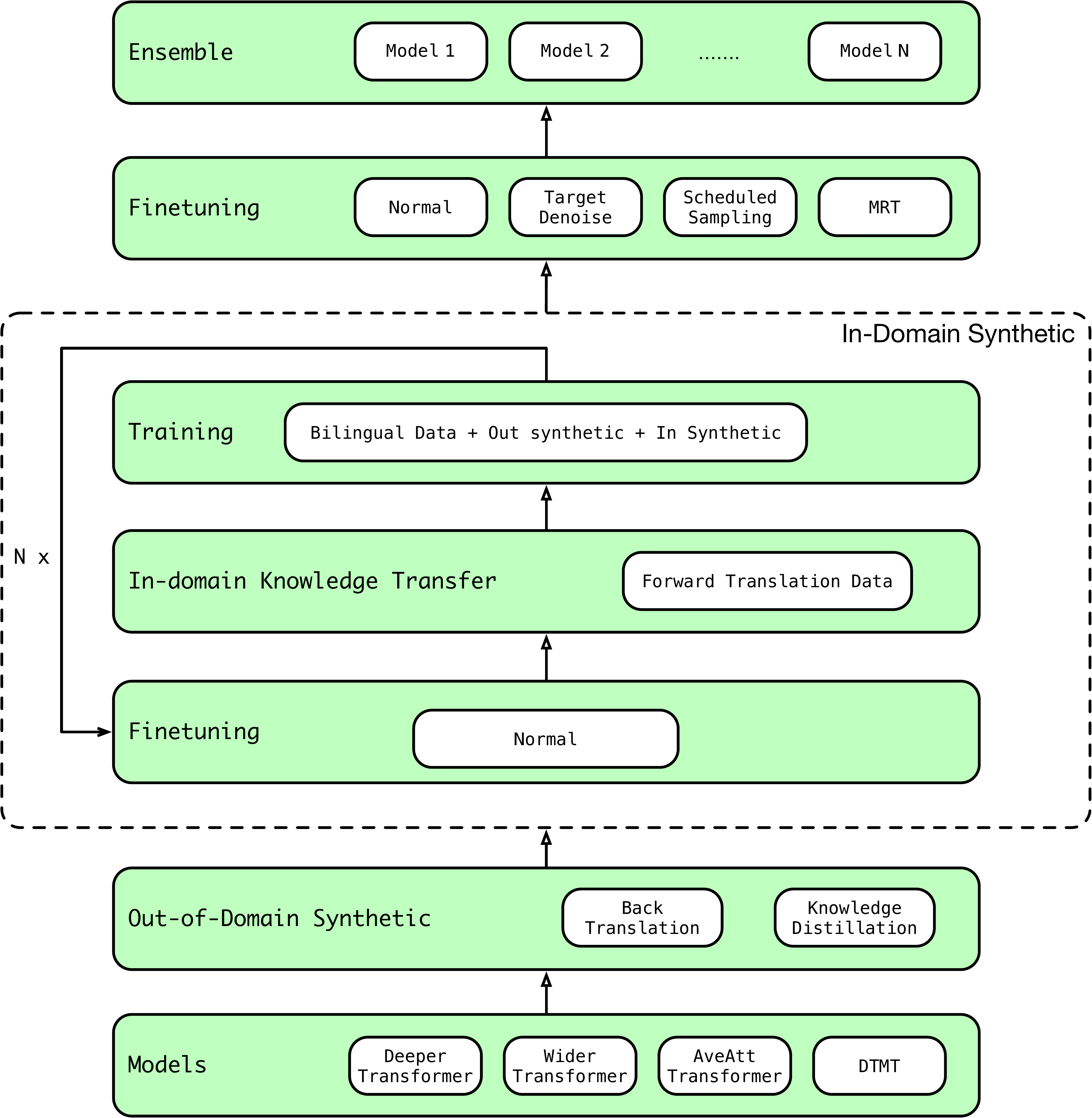}
      \caption{Architecture of WeChat NMT system. For simplicity, the data filtering module is ignored in the overview. } \label{f:main_graph}
 \end{center}
\end{figure*}

\section{System Overview}  \label{techniques}
In this section, we describe our system used in the WMT 2020 news shared task. 

Figure \ref{f:main_graph} depicts the overview of our Wechat NMT. 
Our system can be divided into four parts, namely data filtering, synthetic data generation, in-domain finetuning, and ensemble. The synthetic generation part further includes the generation of out-of-domain and in-domain data. Next, we will illustrate these four parts.


\subsection{Data Filter}
\label{sec:data_filter}
Following previous work~\cite{lietalniutrans2019}, we filter the training bilingual corpus with the following rules:
\begin{itemize}
\item Normalize punctuation with Moses scripts.
\item Filter out the sentences longer than 100 words, or exceed 40 characters in a single word.
\item Filter out the duplicated sentence pairs.
\item The word ratio between the source and the target must not exceed 1:4 or 4:1.
\end{itemize}
We also filter the monolingual corpus with the language model trained by the corresponding data of bilingual training corpus.

\begin{table}[t!]
\centering
\scalebox{0.92} {
\begin{tabular}{l | r}
\hline
 & \bf \textsc{Num}  \\
\hline
Bilingual  Data & 20.7M  \\
Chinese Monolingual Data & 153.5M \\
English Monolingual Data & 121.2M \\
\hline
\end{tabular}
}
\caption{Statistics of all training data.} 
\label{t:train_data}
\end{table}

In our experiments, the bilingual training data is a combination of News Commentary v15, Wiki Titles v2, WikiMatrix, CCMT and the UN corpus. The Chinese monolingual data includes News crawl, News Commentary, Common Crawl and Gigaword corpus. The English monolingual data includes News crawl, News discussions, Europarl v10, News Commentary, Common Crawl, Wiki dumps and the Gigaword corpus. After data filtering, statistics of all training data are shown in Table~\ref{t:train_data}.

\subsection{Out-of-Domain Synthetic Data Generation} 
Now, we describe our techniques for constructing both out-of-domain and in-domain synthetic data. 
The out-of-domain synthetic corpus is generated via both large-scale back-translation and knowledge distillation to enhance the models' performance for all domains. 
Then, we propose iterative in-domain knowledge transfer, which transfers in-domain knowledge to the huge monolingual corpus (i.e., Chinese), and builds our in-domain synthetic corpus.
In the following sections, we elaborate above techniques in detail.

\subsubsection{Large-scale Back-Translation}


Back-translation is shown to be very effective to boost the performance of NMT models in both academic research \cite{hoang2018iterative,edunov2018understanding} and previous years' WMT competitions \cite{deng2018alibaba,baiduwmt2019,ng2019facebook,xia2019microsoft}. Following their work, we also train baseline English-to-Chinese models with the parallel data provided by WMT2020. Both the Left-to-Right Transformer (L2R) and the Right-to-Left Transformer (R2L) are used to translate the filtered monolingual English corpus combined with the English side of golden parallel bitext to Chinese. 
Then the generated Chinese text and the original English text are regarded as the source side and target side, respectively.

In practice, it costs us 7 days on 5 NVIDIA V100 GPU machines to generating all back-translated data.


\subsubsection{Knowledge Distillation}

Knowledge distillation (KD) is proven to be a powerful technique for NMT \cite{kim2016sequence} to transfer knowledge from teacher model to student models.
In particular, we first use the teacher models to generate synthetic corpus in the forward direction (i.e., Chinese$\to$English). Then, we use the generated corpus to train our student models. 

In this work, with baseline Chinese$\to$English models (i.e., L2R and R2L) as teacher models, we translate the Chinese sentences of the parallel corpus to English to form our synthetic KD dataset. The knowledge distillation costs about 2 days on 2 NVIDIA V100 GPU machines to generate all synthetic data.

\subsection{Iterative In-domain Knowledge Transfer}
\label{sec:ind-trans}

Since in-domain finetuning demonstrates substantial BLEU improvements \cite{baiduwmt2019,lietalniutrans2019}, we speculate that the parallel data and the dev/test sets fall in different domains. 
Therefore, adapting our models to the target domain in advance will provide gains over the dev/test sets and give a better initialization point for in-domain finetuning.
To this end, we use knowledge transfer to inject more in-domain information into our synthetic data. 

In particular, we first use normal finetuning (see Section \ref{sec:finetune}) to equip our models with in-domain knowledge.
Then, we ensemble these models and use the ensemble model to translate the Chinese monolingual corpus into English. 
For our ensemble translator, we use 4 models with different architectures.
Next, we pair original Chinese sentences with generated in-domain pseudo English sentences to form a pseudo parallel corpus. 
So far, the in-domain knowledge from ensembled models is transferred to the generated pseudo-parallel corpus. 
Finally, we retrain our model with both the in-domain pseudo-parallel and out-of-domain parallel data.

We refer to the above process as the in-domain knowledge transfer.
In our experiments, we find that iteratively performing the in-domain knowledge transfer can further provide improvements (see Table \ref{t:main_results}). 
For each iteration, we replace the in-domain synthetic data and retrain our models, and it costs about 10 days on 8 NVIDIA V100 GPU machines. 
For the final submission, the knowledge transfer is conducted twice.

\subsection{Data Augmentation}
Aside from synthetic data generation, we also apply two data augmentation methods over our synthetic corpus. 
Firstly, adding synthetic/natural noises to training data is widely applied in the NLP fields \cite{li2017adversarial,belinkov2017synthetic,cheng2019robust} to improve model robustness and enhance model performance. Therefore, we proposed to add token-level synthetic noises. Concretely, we perform random replace, random delete, and random permutation over our data. 
The probability for enabling each of the three operations is 0.1. We refer to this corrupted corpus as \emph{Noisy} data.

Secondly, as illustrated in \citep{edunov2018understanding}, sampling generation over back-translation shows its potential in building robust NMT systems. 
Consequently, we investigate the performance of sampled synthetic data. For back-translated data, we replace beam search with sampling in its generation. 
For in-domain synthetic data, we replace the golden Chinese with the back sampled pseudo Chinese sentences.
We refer to the data with sampling generation as \emph{Sample} data.

As a special case, we refer to the without augmentation data as \emph{Clean} data.

\subsection{In-domain Finetuning}
\label{sec:finetune}

We train the model on large-scale out-of-domain data until convergence and then finetune it on small-scale in-domain data, which is widely used for domain adaption \cite{luong2015stanford,lietalniutrans2019}. Specifically, we take Chinese$\xrightarrow{}$English test sets of WMT 17 and 18 as in-domain data, and filter out documents that are originally created in English \cite{baiduwmt2019}. We name above finetuning approach as \textit{normal} finetuning. In all our finetuning experiments, we set the batch size to 4096, and finetune the model for around 400 steps\footnote{According our experiments, 
finetuing with more steps will make the model easy to overfit on the small in-domain data.} on the in-domain data. 

Furthermore, the well-known problem of exposure bias in sequence-to-sequence generation becomes more serious under domain shift \cite{wangsennrichexposure2020}. To solve this issue, we further explore some \textit{advanced} finetuning approaches and describe details in the following paragraphs. 

\paragraph{Parallel Scheduled Sampling.}
We apply a two-pass decoding strategy for the Transformer decoder when finetuning, which is named as parallel scheduled sampling \cite{mihaylovamartinsscheduled2019,parallel2019schedule}. In the first pass, we obtain model predictions as a standard Transformer, and then mix the predicted sequence with the golden target sequence. In the second pass, we feed above mixture of both golden and predicted tokens as decoder inputs for the final prediction. Thus the 
problem of training-generation discrepancy is alleviated in the finetuning stage.
According to our preliminary experiments, we 
set the proportion of predicted tokens in mixed tokens to 50\%.

\paragraph{Target Denoising.}
In the training stage, the model never sees its own errors. Thus the model trained with teacher-forcing is prone to accumulated errors in testing \cite{ranzato2015exposurebias}. To mitigate this training-generation discrepancy, we add noisy perturbations into decoder inputs when finetuning. Thus the model becomes more robust to prediction errors by target denoising.
Specifically, the finetuning data generator chooses 30\% of sentence pairs to add noise, and keeps remaining 70\% of sentence pairs unchanged. For a chosen pair, we keep the source sentence unchanged, and replace the $i$-th token of target sentence with (1) a random token of current target sentence 15\% of the time (2) the unchanged $i$-th token 85\% of the time.

\paragraph{Minimum Risk Training.}
To further avoid the problem of exposure bias, we propose to use minimum risk training \cite{shenetalmrt2016} in the finetuning stage, which directly optimizes the expected BLEU score instead of the Cross-Entropy loss, and naturally avoids exposure bias. Specifically, the objective is computed by,
\begin{equation}
    R(\theta) = \sum_{s=1}^S \sum_{y \in S(x^{(s)})} Q(y|x^{(s)};\theta, \alpha) \Delta (y, y^{(s)}),
\end{equation}
where $x^{(s)}$ and $y^{(s)}$ are two paired sentences. $\Delta$ denotes a risk function and $S(x^{(s)}) \in Y$ is a sampled subset of full search space. Then, the distribution $Q$ is defined over space $S(x^{(s)})$,
\begin{equation}
    Q(y|x^{(s)};\theta, \alpha) = \frac{P(y|x^{(s}; \theta)^{\alpha}}{\sum_{y'\in \mathcal{S}(x^{(s))}} P(y'|x^{(s}; \theta)^{\alpha}}.
\end{equation}

In practice, we use 4 candidates for each source sentence $x^{(s)}$. Although the paper claimed that sampling generates better candidates, we find that beam search performs better in our extremely large Transformer model. The risk function we used is the 4-gram sentence-level BLEU \cite{chen2014systematic} and we tune the optimal $\alpha$ via grid search within $\{0.005, 0.05, 0.5, 1, 1.5, 2\}$. 
Each model is fine-tuned for a max of 1000 steps.

\begin{table*}[t!]
\centering
\scalebox{0.92} {
\begin{tabular}{l| c| c | c | c}
\hline
\bf \textsc{Settings} & \bf \textsc{Deeper} & \bf \textsc{Wider}  & \bf \textsc{AveAtt} & \bf \textsc{DTMT} \\
\hline
Baseline  & 26.24  & 26.35 & 26.17  & 26.08 \\
+ Back Translation & 29.64  & 29.70 & 29.48  & 28.88 \\
~~~~ + \emph{Finetune} & 35.71  & 35.89 & 35.80 & 35.03 \\
\hline
+ 1st In-domain Knowledge Transfer & 38.14 & 38.22 & 38.21  & 37.98 \\
~~~~ + \emph{Finetune}& 38.36  & 38.25 & 38.13  & 37.85 \\
+ 2nd In-domain Knowledge Transfer & 38.32  & 38.29 & 38.34  & 38.05 \\
~~~~ + \emph{Finetune} & 38.49  & 38.31 &  38.38  & 38.12 \\
~~~~ + \emph{Advanced Finetune} & 39.08  & 39.12 & 38.93  & 38.66 \\
\hline 
+ Normal Ensemble & \multicolumn{4}{c}{39.19}\\
\hline
+ Advanced Ensemble$\star$ & \multicolumn{4}{c}{\textbf{39.89}}\\
\hline
\end{tabular}
}
\caption{Case-sensitive BLEU scores (\%) on the Chinese$\to$English \emph{newstest2019}, where `$\star$' denotes the submitted system.
For each model architecture, we report the highest score among three shards of clean data.
} 
\label{t:main_results}
\end{table*}

\begin{table*}[t!]
\centering
\scalebox{0.92} {
\begin{tabular}{l| c| c | c | c}
\hline
\bf \textsc{Finetuning Approach} & \bf \textsc{Deeper} & \bf \textsc{Wider}  & \bf \textsc{AveAtt} & \bf \textsc{DTMT} \\
\hline
Normal  & 38.49  & 38.31 &  38.38  & 38.12 \\
Parallel Scheduled Sampling & 38.76 & 38.84 & \textbf{38.93} & -- \\
Target Denoising  & 38.88 & 38.92 & 38.63 & \textbf{38.66} \\
Minimum Risk Training & \textbf{39.08} & \textbf{39.12} & 38.78 & 38.45 \\
\hline
\end{tabular}
}
\caption{Case-sensitive BLEU scores (\%) on the Chinese$\to$English \emph{newstest2019} for different finetuning approaches after the 2nd in-domain knowledge transfer. 
For each model architecture, we report the highest score among three shards of clean data and bold the best result among different finetuning approaches.
} 
\label{t:finetune_results}
\end{table*}

\subsection{Ensemble}
\label{sec:ens}
We split each training data into three shards among \textit{Clean}, \textit{Noisy} and \textit{Sample} data respectively, which yields a total number of 9 shards. For each shard, we train seven varieties (two Deeper transformers, two Wider transformers, two AANs and one DTMT) with different model architecture. Then we apply four finetuning approaches on each model, thus the total number of models are quadrupled (about 200 models). 
For ensemble, it is difficult and inefficient to enumerate over all combinations of candidate models (e.g., grid search). Therefore a pruning strategy for model selection is necessary when ensemble. We try to greedily select the top performing models for ensemble. However, only a slight improvement is obtained (less than 0.1 BLEU), as our models are too similar to each other after finetuning.

To further promote diversity among candidate models, we propose the self-bleu driven pruning strategy for \textit{advanced} ensemble. 
Specifically, we take the translation of one model as hypothesis and translations of other models as references. Then we calculate BLEU score for each model to evaluate its diversity among other models. Models with small BLEU scores are selected for ensemble, and vice versa. According to our experiments, we observe that (1) AAN and DTMT show a clear difference with other architectures; (2) data sharding is effective to promote diversity, especially for models trained with \textit{Clean} data; (3) different finetuning approaches cannot bring diversity for the same model. 
Under the guidance of self-bleu scores, our \textit{advanced} ensemble models consists of 20 single models with differences in model architectures, data types, shards and finetuning approaches. As shown in Table \ref{t:main_results}, the \textit{advanced} ensemble achieves absolute improvements over the normal ensemble (up to 0.7 BLEU improvements). 


\section{Experiments}  \label{experiments}
\subsection{Settings}
All of our experiments are carried out on 15 machines with 8 NVIDIA V100 GPUs each of which have 32 GB memory. 
We use cased BLEU scores calculated with Moses\footnote{http://www.statmt.org/moses/} mteval-v13a.pl script as evaluation metric. 
\emph{newstest2019} is used as the development set.
For all experiments, we use LazyAdam optimizer with $\beta_{1}$ = 0.9, $\beta_{2}$ = 0.998 and $\epsilon=10^{-9}$. The learning rate is set to $2.0$ and decay with training steps. We use warmup step = 8000. We set beam size to 4 and alpha to 0.6 during decoding.

\subsection{Pre-processing and Post-processing}
We segment the Chinese sentences with an in-house word segmentation tool. 
For English sentences, we successively apply punctuation normalization, tokenization and truecasing with the scripts provided in Moses.
To enable open-vocabulary, we use byte pair encoding BPE~\cite{sennrichetal2016bpe} with 32K operations for both Chinese and English sides.

For the post-processing, we apply de-truecaseing and de-tokenizing on the English translations with the scripts provided in Moses.

\subsection{Main Results}
Table \ref{t:main_results} shows that the translation quality is largely improved with proposed techniques. 
We observe a solid improvement of 2.8$\sim$3.4 BLEU for the baseline system after back translation.
In-domain finetuning yields substantial improvements among all model architectures, which are 6.07$\sim$6.32 BLEU. 
The finetuned Transformer models achieve about 35.89 BLEU scores, and the DTMT achieves a 35.03 BLEU score. These findings demonstrate that the domain of training corpus is apart from the target domain, and hence domain adaptation has great potential in improving model performance in the target domain.

As described in Section \ref{sec:ind-trans}, we inject the in-domain knowledge into our monolingual corpus. Two In-domain knowledge transfers provide another up to 3.02 BLEU score gain (i.e., from about 35.03 to 38.05). The in-domain knowledge transfer brings more improvement compared with the normal finetuned models. 
Besides, we find that models further finetuned after in-domain transfer performs slightly better (about 0.1 BLEU). The improvement suggests that although in-domain transfer has already provided plenty of in-domain knowledge, it still has room for in-domain finetuning. 
We further apply \textit{advanced} finetuning techniques to our models, as described in Section \ref{sec:finetune}. The advanced finetuning further brings about 0.81 BLEU score gains, and we obtain our best single model with 39.12 BLEU scores. 
 
In our preliminary ensemble experiments, we combine some top performing models at each decoding step, but only achieve slight improvement over single models (about 0.1 BLEU). 
With our \textit{advanced} ensemble strategies in section \ref{sec:ens}, further improvements are achieved over the normal ensemble (0.7 BLEU). As a result, our WMT 2020 Chinese$\to$English submission achieves a cased BLEU score of 36.9 on \emph{newstest2020}, which is the highest among all submissions.

\subsection{Effects of Advanced Finetuning Approaches}
In this section, we describe our experiments on advanced finetuning. Here we take clean models as examples, but models trained with noisy data and sampled data show similar trends.

As shown in Table \ref{t:finetune_results}, all three advanced finetuning methods significantly outperform normal finetuning. For Wider and Deeper Transformers, Minimum Risk Training provides the highest BLEU gain, which is 0.81. For Average Attention Transformer, Parallel Schedule Sampling improves the model performance from 38.38 to 38.93. For the DTMT model, Target Denoising performs the best, improving from 38.12 to 38.66. 
These findings are in line with the conclusion of \citet{wangsennrichexposure2020} that links exposure bias with domain shift. 
For each type of model, we only keep the best-performing finetuned one for the final model ensemble.

\section{Conclusion}  \label{conclusion}
In this paper, we introduce the system WeChat submitted for the WMT 2020 shared task on Chinese$\to$English news translation. Our system is based on the Transformer~\cite{transformer2017} with different variants and the DTMT~\cite{meng2019dtmt} architecture. Data selection, several effective synthetic data generation approaches (i.e., back-translation, knowledge distillation, and iterative in-domain knowledge transfer), advanced finetuning approaches (i.e., parallel scheduled sampling, target denoising, and minimum risk training) and self-bleu based model ensemble are employed and proven effective in our experiments. Our constrained Chinese$\to$English system achieved 36.9 case-sensitive BLEU score which is the highest among all submissions.

\bibliography{emnlp2020}
\bibliographystyle{acl_natbib}

\end{document}